\documentclass[journal,12pt,onecolumn,draftclsnofoot,]{IEEEtran}
\usepackage[left=1.97cm, right=1.97cm, top=2.25cm, bottom=2.7cm, includefoot]{geometry}

\pdfpageattr{/Group << /S /Transparency /I true /CS /DeviceRGB >>}

\usepackage[utf8]{inputenc}
\usepackage[T1]{fontenc}
\usepackage{textcomp}
\usepackage{microtype}
\usepackage{calc}
\usepackage[normalem]{ulem}
\usepackage{verbatim}

\usepackage{lipsum}

\usepackage[range-phrase=--,per-mode=symbol-or-fraction,range-units=single,list-units=single,detect-all,forbid-literal-units=true]{siunitx}
\AtBeginDocument{
\DeclareSIUnit\bit{bit}
\DeclareSIUnit\byte{Byte}
\DeclareSIUnit\mbps{\mega\bit\per\second}
\DeclareSIUnit\kmh{\kilo\meter\per\hour}
\DeclareSIUnit\mw{\milli\watt}
\DeclareSIUnit\decibelm{dBm}
\DeclareSIUnit\decibeli{dBi}
\DeclareSIUnit\vehicle{veh}
}

\usepackage{silence}
\usepackage{csquotes}
\usepackage[backend=biber,style=ieee,doi=false,isbn=false,mincitenames=1,maxcitenames=2]{biblatex}
\DeclareFieldFormat{sentencecase}{#1} % never apply sentence casing, even if bibtex field is unprotected
\DeclareFieldFormat{titlecase}{#1} % never apply title casing, even if bibtex field is unprotected
\usepackage{xpatch}
\xpatchbibmacro{textcite}{\addspace}{\addnbspace}{}{}
\xpatchbibmacro{Textcite}{\addspace}{\addnbspace}{}{}
\DefineBibliographyStrings{english}{
	andothers = et~al\adddot\addspace
}

\DeclareSourcemap{
	\maps[datatype=bibtex]{
		\map[overwrite=true]{
			\step[fieldsource=school,fieldtarget=institution]
		}
	}
}

\usepackage{amsmath}

\usepackage{amssymb}
\usepackage{amsfonts}
\usepackage{amsthm}
\usepackage{algpseudocode}

\usepackage{booktabs}
\usepackage{url}

\usepackage[inline]{enumitem}

\usepackage[caption=false,font=footnotesize]{subfig}
\usepackage[pdftex]{graphicx}
\DeclareGraphicsExtensions{.pdf,.png,.jpg}
\usepackage{tikz}
\usetikzlibrary{arrows}
\usetikzlibrary{calc}
\usetikzlibrary{chains}
\usetikzlibrary{scopes}
\usepackage[american]{babel}
\usepackage{hyphenat}

\usepackage[capitalize,noabbrev,nameinlink]{cleveref}

\RequirePackage{xstring}
\RequirePackage{xparse}
\RequirePackage[]{acro}
\makeatletter
\@ifpackagelater{acro}{2019/02/17}{
	\NewDocumentCommand\acrodef{mO{#1}mG{}}{\DeclareAcronym{#1}{short={#2}, long={#3}, foreign-plural={}, #4}}
}{
	\NewDocumentCommand\acrodef{mO{#1}mG{}}{\DeclareAcronym{#1}{short={#2}, long={#3}, #4}}
}
\makeatother

\acrodef{AI}{Artificial Intelligence}
\acrodef{BER}{Bit Error Rate}
\acrodef{CSI}{Channel State Information}
\acrodef{IoT}{Internet of Things}
\acrodef{LBT}{Listen-Before-Talk}
\acrodef{MCS}{Modulation and Coding Scheme}
\acrodef{NG-TCMS}{Next-Generation Train Control and Monitoring System}
\acrodef{NR}{New Radio}
\acrodef{QoS}{Quality of Service}
\acrodef{SCI}{Sidelink Control Information}
\acrodef{SCS}{Subcarrier Spacing}
\acrodef{SL}{Sidelink}
\acrodef{SNR}{Signal-to-Noise-Ratio}
\acrodef{SPS}{Semi-Persistent Scheduling}
\acrodef{RSRP}{Reference Signal Received Power}
\acrodef{TB}{Transport Block}
\acrodef{TSN}{Time-Sensitive Networking}
\acrodef{UE}{User Equipment}
\acrodef{V2I}{Vehicle-to-Infrastructure}
\acrodef{V2P}{Vehicle-to-Pedestrian}
\acrodef{V2V}{Vehicle-to-Vehicle}
\acrodef{V2X}{Vehicle-to-Everything}
\acrodef{WLCN}{WireLess Consist Network}
\acrodef{WLTB}{WireLess Train Backbone}
\acrodef{PRR}{Packet Reception Ratio}
\acrodef{PRB}{Physical Resource Block}
\acrodef{SINR}{Signal-to-Interference-plus-Noise-Ratio}
\acrodef{gNB}{g-NodeB}
\acrodef{LoS}{Line-of-Sight}

\acrodef{B.A.T.M.A.N.}{Better Approach To Mobile Ad-hoc Networking}
\acrodef{TTL}{Time to Live}
\acrodef{RB}{Resource Block}
\acrodef{OGM}{Originator Message}
\acrodef{MPR}{Multipoint Relaying}
\acrodef{PDR}{Packet delivery ratio}
\acrodef{D2D}{Device-to-Device}
\acrodef{OLSR}{Optimized Link State Routing Protocol}
\acrodef{AODV}{Ad hoc On-Demand Distance Vector Protocol}

\usepackage{todonotes}

\def\todoCtd#1{%
	TODO: #1%
	\ifx&#1&...\fi%
	\endgroup
	\relax
}

\NewDocumentCommand\IEEE{ s m >{\SplitArgument{4}{/}}d[] }{%
	\IfBooleanTF{#1}{}{IEEE\,}% suppress IEEE when using starred form
	\nolinebreak[2]% this is a somewhat bad place for a line break
	#2%
	\IfNoValueTF{#3}{%
	}{%
		\sommerIEEELettersSlashed#3%
	}%
}
\newcommand{\sommerIEEELettersSlashed}[5]{%
	\IfNoValueTF{#2}{%
	}{%
		\nolinebreak[3]% multiple letters, this is just a very bad place for a line break
	}%
	#1%
	\IfNoValueTF{#2}{}{/#2}%
	\IfNoValueTF{#3}{}{/#3}%
	\IfNoValueTF{#4}{}{/#4}%
	\IfNoValueTF{#5}{}{/#5}%
}

\usepackage{pifont}

\begin{document}

\title{Toward a Graph-Theoretic Model of Belief:\\ Confidence, Credibility, and Structural Coherence}

\author{%
\IEEEauthorblockN{%
    Saleh Nikooroo
}%

\small{
    \texttt{%
	saleh.nikooroo%
	@gmail.com%
    }
}
\\
}

\maketitle

%\begin{abstract}
%Despite their impressive performance, contemporary neural networks often lack structural safeguards that promote stable learning and interpretable behavior. In this work, we introduce a reformulation of layer-level transformations that departs from the standard unconstrained affine paradigm. Each transformation is decomposed into a structured linear operator and a residual corrective component, enabling more disciplined signal propagation and improved training dynamics.
%Our formulation encourages internal consistency and supports stable information flow across depth, while remaining fully compatible with standard learning objectives and backpropagation. Through a series of synthetic and real-world experiments, we demonstrate that models constructed with these structured transformations exhibit improved gradient conditioning, reduced sensitivity to perturbations, and layer-wise robustness. We further show that these benefits persist across architectural scales and training regimes. This study serves as a foundation for a more principled class of neural architectures that prioritize stability and transparency—offering new tools for reasoning about learning behavior without sacrificing expressive power.
%\end{abstract}

\begin{abstract}
Belief systems are often treated as globally consistent sets of propositions or as scalar-valued probability distributions. Such representations tend to obscure the internal structure of belief, conflate external credibility with internal coherence, and preclude the modeling of fragmented or contradictory epistemic states. This paper introduces a minimal formalism for belief systems as directed, weighted graphs. In this framework, nodes represent individual beliefs, edges encode epistemic relationships (e.g., support or contradiction), and two distinct functions assign each belief a \emph{credibility} (reflecting source trust) and a \emph{confidence} (derived from internal structural support). Unlike classical probabilistic models, our approach does not assume prior coherence or require belief updating. Unlike logical and argumentation-based frameworks, it supports fine-grained structural representation without committing to binary justification status or deductive closure. The model is purely static and deliberately excludes inference or revision procedures. Its aim is to provide a foundational substrate for analyzing the internal organization of belief systems, including coherence conditions, epistemic tensions, and representational limits. By distinguishing belief structure from belief strength, this formalism enables a richer classification of epistemic states than existing probabilistic, logical, or argumentation-based approaches.
\end{abstract}

\begin{IEEEkeywords}
Belief systems, epistemic graphs, confidence, credibility, structural coherence, epistemology, knowledge representation.
\end{IEEEkeywords}

\acresetall
\IEEEpeerreviewmaketitle

% -------------- Section end marker --------------
%                _       _
%               ( )_    ( )
%    ___  _   _ | ,_)   | |__     __   _ __   __
%  /'___)( ) ( )| |     |  _ `\ /'__`\( '__)/'__`\
% ( (___ | (_) || |_    | | | |(  ___/| |  (  ___/
% `\____)`\___/'`\__)   (_) (_)`\____)(_)  `\____)
%
% -------------- Section end marker --------------

% USEFULE EXAMPLES:
%\SI{6.2}{\mega\bit\per\second}
%\SI{2.4}{\giga\hertz}
%\SI{50}{\kilo\meter\per\hour}
%\SIlist{10;20;40;50}{\kilo\meter\per\hour}
%\SIrange{10}{50}{\kilo\meter\per\hour}
%\cref{introd}

\section{Introduction}
\label{intro}

Belief systems play a central role in both human cognition and artificial intelligence. They encode an agent's understanding of the world, provide the basis for decision-making, and often serve as the implicit substrate for reasoning and inference. Despite their importance, belief systems are frequently modeled in oversimplified ways: as unordered sets of propositions, scalar-valued probability distributions, or deductively closed and globally consistent knowledge bases.

Such representations obscure key structural properties of belief—such as how individual beliefs support, contradict, or qualify one another—and often conflate distinct epistemic concepts like source reliability and internal coherence. Classical logical approaches emphasize syntactic manipulation and consistency but exclude partial incoherence by design. Probabilistic models reduce belief to scalar values and focus on inference, sidelining the topology of epistemic relationships. Belief revision frameworks such as AGM offer mechanisms for updating beliefs but operate under rationality postulates that prevent representation of fragmented or internally contradictory belief states. Even recent argumentation-based models, while structurally richer, typically collapse belief status to discrete logical outcomes (accepted, rejected, undecided) and do not explicitly model degrees of confidence or the separation between external trust and internal justification.

This paper proposes a structural foundation for belief systems using directed, weighted graphs. In this framework, each node represents an individual belief, while edges encode epistemic relationships—such as support, contradiction, or qualification—between beliefs. Crucially, each belief is endowed with two distinct attributes: \emph{credibility}, which reflects external source reliability, and \emph{confidence}, which captures internal structural support within the belief graph. This dual representation enables the modeling of belief systems that are internally fragmented, non-monotonic, or epistemically inconsistent, yet locally coherent.

Our goal is not to define a new inference calculus or belief update procedure. Instead, we offer a minimal, static formalism that serves as a substrate for representing and analyzing the internal organization of belief systems. By separating belief structure from belief strength, and by supporting fine-grained representation of epistemic support and tension, the proposed model enables a broader and more flexible analysis than standard probabilistic, logical, or argumentation-based paradigms.

The remainder of the paper is organized as follows. Section~\ref{sec:related} surveys classical and contemporary models of belief representation, identifying their core assumptions and structural limitations. Section~\ref{sec:formal} introduces our graph-based formalism. Section~\ref{sec:coherence} defines structural coherence and characterizes incoherent configurations. Section~\ref{sec:credconf} formalizes the distinction between credibility and confidence. Section~\ref{sec:usecases} outlines intended applications of the model. We conclude in Section~\ref{sec:conclusion} with a discussion of potential extensions and design choices deliberately deferred.

% -------------- Section end marker --------------
%                _       _
%               ( )_    ( )
%    ___  _   _ | ,_)   | |__     __   _ __   __
%  /'___)( ) ( )| |     |  _ `\ /'__`\( '__)/'__`\
% ( (___ | (_) || |_    | | | |(  ___/| |  (  ___/
% `\____)`\___/'`\__)   (_) (_)`\____)(_)  `\____)
%
% -------------- Section end marker --------------

\section{Related Work}
\label{sec:related}

Belief systems have long been central to inquiry in philosophy, formal logic, artificial intelligence, and epistemology. Each tradition has contributed distinct modeling paradigms—from probabilistic inference to modal logic and argumentation theory. Yet despite their technical depth, these approaches often rest on foundational assumptions that limit their ability to represent belief systems with internal structure, epistemic fragmentation, or contradictory subcomponents. Most models focus on belief content, inference validity, or revision dynamics, rather than on the architecture of belief interdependence or the separation of epistemic roles. This section begins by identifying core modeling assumptions shared across traditions, then systematically contrasts our approach with representative frameworks.

\subsection*{Motivations and Modeling Gaps}

Three structural limitations recur across classical and contemporary models:

\begin{itemize}
\item \textbf{Assumption of global coherence}: Belief systems are presumed to be logically consistent, or inconsistencies are treated as noise to be eliminated via revision.
\item \textbf{Flat or atomistic representations}: Beliefs are typically modeled as isolated propositions, modal statements, or probabilistic assignments, without encoding relational structure such as qualification, contradiction, or support.
\item \textbf{Unidimensional belief strength}: A single numerical or logical measure (e.g., probability, certainty, truth value) is used to quantify belief strength, thereby conflating external credibility with internal coherence.
\end{itemize}

These assumptions constrain the expressiveness of belief models in domains characterized by epistemic tension, layered justification, or incomplete reconciliation. In contrast, our approach introduces a static, graph-based formalism that makes no commitment to inference, truth-functional semantics, or revision mechanisms. It explicitly separates externally assigned \emph{credibility} from internally derived \emph{confidence}, and models structural coherence directly through graph topology. This enables the representation and analysis of fragmented, tension-laden, or partially coherent belief systems that fall outside the representational scope of traditional paradigms.

\subsection*{Belief as Probability: Bayesian and Probabilistic Models}

Bayesian models represent belief as a scalar probability assigned to propositions and updated via Bayes’ rule. These frameworks are powerful for inference and decision-making under uncertainty but presuppose global coherence and a single quantitative measure of belief strength. Graphical extensions like Bayesian networks~\cite{pearl1988probabilistic} encode conditional dependencies among variables but do not model epistemic tension, partial justification, or internal fragmentation. They lack a notion of belief structure beyond probabilistic dependence and do not distinguish between source-derived trust and internally emergent support. As such, they are ill-suited for analyzing the internal organization or coherence of belief systems in static or non-inferential settings.

\subsection*{Belief Logic: Modal and Doxastic Formalisms}

Modal logics of belief, including doxastic systems~\cite{liang2022epistemic}, provide formal tools to reason about belief introspection, consistency, and knowledge attribution. These logics typically use operators like $\Box p$ to express belief in $p$ and support nesting, introspection, and agent-relative perspectives. However, they represent beliefs syntactically and presuppose logical closure and idealized rationality. There is no capacity to model internal epistemic structure—such as how one belief qualifies or contradicts another—nor to capture fragmented or partially coherent belief systems. The assumption of global consistency and the absence of structural relationships among beliefs make these logics incompatible with the kind of graph-based, coherence-sensitive representation developed in our framework.

\subsection*{Belief Revision: AGM Theory and Extensions}

The AGM framework~\cite{agm1985} formalizes rational belief change through operations like expansion, revision, and contraction, grounded in set-theoretic representations. While it addresses the integration of new information and preservation of consistency, it treats belief systems as deductively closed sets of propositions and does not provide mechanisms for representing epistemic structure or internal relational support. Moreover, it conflates belief justification with truth-preserving revision and does not account for degrees of source credibility or internally constructed confidence. In contrast, our model allows belief systems to be structurally incoherent or fragmented, and explicitly separates external trust from internal coherence without assuming inference or closure.

\subsection*{Argumentation Frameworks and Epistemic Graphs}

Argumentation-based models, such as Dung’s abstract framework~\cite{dung1995}, represent beliefs or claims as nodes connected by attack or support relations. Recent extensions, including epistemic graphs~\cite{hunter2020}, add expressiveness by incorporating belief annotations or weighted valuations, and tools like SAT-based solvers have been applied to model belief propagation~\cite{hunter2022epistemic}. These models bring structural representations closer to our graph-based approach, especially in emphasizing inter-argument relationships. However, their core focus remains on determining the acceptability of arguments under logical constraints, often aiming to resolve disputes through computational adjudication. Work by Huang et al.~\cite{huang2023belief} and Idlemassoud et al.~\cite{idlemassoud2022qualitative} introduces belief-based support structures, but similarly prioritizes inferential resolution or consistency over representing epistemic fragmentation. In contrast, our framework foregrounds representation over inference and is designed to preserve internal tension and unresolved conflict as legitimate structural features.

\subsection*{Graph-Theoretic Models of Cognitive and Social Beliefs}

Graphical approaches to social and cognitive belief modeling—such as the BENDING framework~\cite{vlasc2023bending}—emphasize belief dynamics under environmental pressures, modeling how informational inputs shape multi-dimensional belief landscapes. Other works, like that of Camina et al.~\cite{carmina2022hubs}, explore the structural roles of belief ``hubs'' in group belief systems, revealing sociocultural patterns in belief formation. These approaches are valuable for domain-specific modeling, particularly in cognitive science and social epistemology. However, they are typically tied to empirical data and dynamic behavior, often requiring domain assumptions or time-evolution models. Our formalism, by contrast, is domain-agnostic and static, focusing on the intrinsic structure of belief systems rather than their dynamics or behavioral evolution.

\subsection*{Logical and Probabilistic Representations}

Alternative models that transcend classical logic and probability have also been proposed. Dubois~\cite{dubois2022representing} surveys such efforts, noting attempts to expand the expressiveness of belief representation while preserving inferential soundness. Schwind~\cite{schwind2022darwiche} offers a treatment of iterated revision via ordinal conditional functions, enhancing the modeling of belief updates over time. These frameworks, while more nuanced than basic probabilistic or modal systems, still prioritize reasoning accuracy and inferential tractability. Our contribution diverges from this lineage by explicitly avoiding inferential mechanisms, instead prioritizing structural transparency and the representation of internal epistemic relationships, including unresolved contradictions or partial support.

\subsection*{Structural and Epistemic Foundations}

Several recent works have emphasized the importance of structure in belief modeling. Baltag~\cite{baltag2022justified} develops topological semantics for modeling knowledge, belief, and evidence, which offers conceptual insight into layered coherence. Bizyaeva et al.~\cite{bizyaeva2023multi} simulate belief bifurcations in dynamic social networks, while Souza~\cite{souza2022hyperintensional} models fine-grained epistemic change via hyperintensional logics. While each of these frameworks acknowledges complexity in belief systems, they are often tied to dynamic processes, semantic interpretations, or specific logical foundations. In contrast, our model strips down to a minimal static structure that cleanly separates source-based \emph{credibility} from structure-derived \emph{confidence}, supporting analysis of epistemic fragmentation without reliance on truth-functional semantics or dynamic procedures. Other structurally focused works, such as Jirousek et al.~\cite{jirousek2023graphical} and Douven~\cite{douven2022bounded}, provide important tools for understanding local belief dynamics and support aggregation, yet still do not model internal epistemic architecture as our graph-based formalism does. Santos et al.~\cite{santos2022iterative} offer iterative belief consolidation from peer networks—a direction compatible with possible extensions of our confidence function in future work.

\subsection*{Neural and Applied Models}

Recent advances in neural modeling have begun to recognize the importance of structured belief representations. Hase et al.~\cite{hase2023methods} introduce SLAG, a belief-consistency regularization framework that aligns language model outputs with graph-structured belief representations, illustrating the role of structure in improving coherence in AI-generated content. Olsson~\cite{olsson2024analogy} explores analogical reasoning as a structural alignment problem, suggesting that belief similarity may be grounded in topological correspondence rather than content alone. Burgueno et al.~\cite{burgueno2022belief} address the practical challenge of integrating conflicting stakeholder beliefs into unified domain models under uncertainty, but rely on procedural reconciliation rather than structural preservation of divergence. Our approach diverges by prioritizing representation over resolution: instead of enforcing agreement, it provides a substrate for analyzing tension, redundancy, or fragmentation without enforcing convergence.

\subsection*{Dynamic and Epistemic Network Models}

A growing body of work models belief evolution over time using dynamic networks. Tenorio et al.~\cite{tenorio2024tracking} apply state-space techniques to model temporal changes in belief networks. Michalski~\cite{michalski2022temporal} simulates consensus formation in dynamic epistemic graphs, while Li et al.~\cite{li2024optimization} optimize the structure of belief-driven clusters. Bernardo~\cite{bernardo2024bounded} surveys bounded confidence models, where belief change is constrained by epistemic distance—a notion reminiscent of our definition of local structural coherence. However, these models presuppose update dynamics and convergence goals. In contrast, our static formalism is deliberately agnostic to temporal processes, designed to make belief structure analyzable independent of dynamics, while still enabling future extensions in that direction.

\subsection*{Probabilistic Reasoning and Graphical Inference}

Many frameworks focus on enhancing probabilistic reasoning through structural or logical extensions. Rocha~\cite{Rocha2022} combines L-stable and credal semantics to enrich probabilistic logic programming. Jaeger~\cite{Jaeger2023a,Jaeger2022} surveys the intersection of probabilistic inference and graph-based reasoning, emphasizing learning-based integration. Wang~\cite{Wang2022} proposes TRUST, a framework that learns belief network structures under uncertainty. Connors~\cite{Connors2022} bridges cognitive neuroscience and belief dynamics, while Freedman~\cite{Freedman2023} uses Bayesian tools to infer beliefs from knowledge graphs. Bílková~\cite{Bilkova2022} offers a qualitative logic-based model for belief uncertainty. While powerful in their inferential capabilities, these systems typically assume that structure exists to serve reasoning. Our model, by contrast, treats structure as primary: inference is not presupposed, allowing for epistemic states that are unresolved, inconsistent, or only partially connected.

\subsection*{Distinctiveness of the Present Approach}

Despite the richness of existing models, none to our knowledge provide a static, graph-based formalism with the following properties:

\begin{itemize}
    \item \textbf{Separation of credibility and confidence}: Unlike probabilistic or logical models that collapse belief strength into a single scalar, we differentiate external source trust (credibility) from internal structural support (confidence).
    \item \textbf{Structural coherence}: Our model enables reasoning about coherence, fragmentation, and internal tension without requiring logical closure or truth-functional interpretation.
    \item \textbf{Domain-agnostic formulation}: The formalism abstracts away from specific knowledge domains, enabling applicability to a wide range of epistemic contexts without domain-specific assumptions.
    \item \textbf{Non-deductive design}: The model is not tied to inferential mechanisms or update procedures; it captures belief architecture independently of truth, deduction, or revision.
\end{itemize}

This structural substrate is intended not to replace inferential or dynamic models, but to complement them—offering a foundational representation of epistemic configuration on which reasoning, learning, or update mechanisms may later operate.

% -------------- Section end marker --------------
%                _       _
%               ( )_    ( )
%    ___  _   _ | ,_)   | |__     __   _ __   __
%  /'___)( ) ( )| |     |  _ `\ /'__`\( '__)/'__`\
% ( (___ | (_) || |_    | | | |(  ___/| |  (  ___/
% `\____)`\___/'`\__)   (_) (_)`\____)(_)  `\____)
%
% -------------- Section end marker --------------
\section{Formal Structure of a Belief System}
\label{sec:formal}

We introduce a static formalism for belief systems that explicitly distinguishes structural organization from inferential behavior. Unlike probabilistic or logical systems that treat belief as scalar or syntactic, our framework models belief systems as weighted directed graphs with heterogeneous relational semantics and dual epistemic evaluations. This design enables representation of fragmented, incoherent, or non-resolved belief configurations that remain opaque to traditional models.

\subsection*{Definition 1 (Belief System)}
A belief system is defined as a quadruple
\[
B = (N, E, \mathrm{cred}, \mathrm{conf})
\]
where:
\begin{itemize}
    \item $N$ is a finite set of nodes, each representing a distinct belief, proposition, or assumption.
    \item $E \subseteq N \times N$ is a set of directed edges denoting epistemic relationships between beliefs.
    \item $\mathrm{cred}: N \rightarrow [0,1]$ assigns a \emph{credibility score} to each belief, capturing external trust in its source.
    \item $\mathrm{conf}: N \rightarrow [0,1]$ assigns a \emph{confidence score}, reflecting internal support derived from the system's structural configuration.
\end{itemize}

This dual attribution decouples belief origin from structural integration—an expressivity gap left unaddressed in classical models such as AGM theory, probabilistic logic, and modal belief systems.

\subsection*{Edge Semantics and Types}
Each directed edge $(n_i, n_j) \in E$ encodes an epistemic relation from belief $n_i$ to belief $n_j$. To enrich this structure, we define an optional edge-labeling function:
\[
\mathrm{type}: E \rightarrow \{\text{support}, \text{qualification}, \text{contradiction}\}
\]
enabling semantic differentiation among relational types:

\begin{itemize}
    \item \textbf{Support}: $n_i$ justifies or reinforces $n_j$.
    \item \textbf{Qualification}: $n_i$ contextualizes or nuances $n_j$.
    \item \textbf{Contradiction}: $n_i$ undermines or negates $n_j$.
\end{itemize}

While argumentation frameworks~\cite{dung1995, hunter2020} typically model binary attack/support with fixed semantics, our approach introduces a more flexible and extensible relational vocabulary. Importantly, our model does not require edge labels to be used in formal operations—they are optional and can be ignored for minimal deployment.

\subsection*{Static Structure and Flexibility}

The graph $B$ is treated as a static, domain-agnostic object. We intentionally do not define belief update procedures, dynamic learning mechanisms, or inference strategies. Unlike belief networks~\cite{pearl1988probabilistic} or iterated revision frameworks~\cite{schwind2022darwiche}, our model does not assume acyclicity, consistency, or deductive closure.

This minimalism allows the representation of belief systems that are:
\begin{itemize}
    \item internally fragmented (disconnected or modular),
    \item epistemically incoherent (containing contradictions), or
    \item structurally ambiguous (multiple overlapping supports or unresolved tensions).
\end{itemize}

Such configurations remain difficult to capture in existing frameworks that impose rationality postulates or inferential completeness.

\subsection*{Interpretive Constraint and Minimal Commitment}

Each node $n \in N$ is assumed to carry propositional content, though the nature of that content (e.g., syntactic, semantic, modal) is not specified. The functions $\mathrm{cred}$ and $\mathrm{conf}$ are externally assigned and need not be derived from any inference mechanism. This explicit commitment to static, non-deductive modeling contrasts with Bayesian, modal, or logical systems that tightly couple structure to belief propagation or truth maintenance.

\vspace{0.5em}
\noindent In summary, this formal structure provides a low-assumption yet highly expressive substrate for analyzing belief organization and coherence, intentionally separating structure from truth, credibility from confidence, and representation from reasoning. It fills a modeling niche between logical rigor and empirical flexibility, suitable for systems that must handle partial, conflicting, or source-diverse beliefs.

% -------------- Section end marker --------------
%                _       _
%               ( )_    ( )
%    ___  _   _ | ,_)   | |__     __   _ __   __
%  /'___)( ) ( )| |     |  _ `\ /'__`\( '__)/'__`\
% ( (___ | (_) || |_    | | | |(  ___/| |  (  ___/
% `\____)`\___/'`\__)   (_) (_)`\____)(_)  `\____)
%
% -------------- Section end marker --------------

% -------------- Section end marker --------------
%                _       _
%               ( )_    ( )
%    ___  _   _ | ,_)   | |__     __   _ __   __
%  /'___)( ) ( )| |     |  _ `\ /'__`\( '__)/'__`\
% ( (___ | (_) || |_    | | | |(  ___/| |  (  ___/
% `\____)`\___/'`\__)   (_) (_)`\____)(_)  `\____)
%
% -------------- Section end marker --------------

\section{Coherence and Incoherence in Belief Structures}
\label{sec:coherence}

The notion of coherence in a belief system has traditionally been tied to logical consistency, deductive closure, or probabilistic convergence. In contrast, we define coherence as a structural property—independent of inference mechanisms or truth-functional semantics. Our model supports both localized and systemic coherence assessments based on edge types and belief interdependencies, without presupposing rationality postulates or resolution mechanisms.

\subsection*{Definition 2 (Local Coherence)}
Let $B = (N, E, \mathrm{cred}, \mathrm{conf})$ be a belief system. A subset $S \subseteq N$ is said to be \emph{locally coherent} if:
\begin{enumerate}
    \item There exists no pair $n_i, n_j \in S$ such that $(n_i, n_j) \in E$ and $\mathrm{type}(n_i, n_j) = \text{contradiction}$.
    \item The subgraph induced by $S$ contains only \texttt{support} or \texttt{qualification} edges.
\end{enumerate}

This generalization of consistency allows for fragmented zones of coherence within a globally incoherent belief graph. In contrast to modal and probabilistic models, which often assume system-wide consistency, our formalism permits epistemic heterogeneity across the graph.

\subsection*{Definition 3 (Global Coherence)}
A belief system $B$ is \emph{globally coherent} if all edges preserve epistemic compatibility—that is, the graph contains no path or cycle along which contradictions are propagated, and no nodes are epistemically undermined by upstream beliefs.

This is a strong structural condition not assumed by default. Rather than treating incoherence as a flaw, we treat global coherence as an exceptional state that may be absent from real-world systems characterized by contradiction, conflict, or underdetermination.

\subsection*{Definition 4 (Structural Incoherence)}
We define \emph{structural incoherence} in two general forms:
\begin{itemize}
    \item \textbf{Contradictory Subgraphs}: The presence of cycles or chains of epistemic contradiction.
    \item \textbf{Undersupported Beliefs}: Nodes whose confidence scores are high despite receiving no or only incoherent support within the graph.
\end{itemize}

This captures types of structural instability typically unrecognized in probabilistic belief networks or deductive logics, which either suppress or resolve such inconsistencies by design.

\subsection*{Coherence Relative to Confidence}
Our model allows but does not require an interpretive alignment between assigned confidence values and graph topology. This leads to an evaluative notion:

\begin{quote}
A belief system $B$ is \emph{confidence-consistent} if (i) no belief with high $\mathrm{conf}$ is structurally undermined via contradiction paths, and (ii) no belief with low $\mathrm{conf}$ receives strong, consistent support from highly confident sources.
\end{quote}

This notion is diagnostic, not prescriptive. Unlike belief propagation in Bayesian or neural systems~\cite{hase2023methods}, confidence in our model is not inferred but supplied externally, allowing users to highlight epistemic misalignment or unjustified certainties.

\subsection*{Remarks on Structural Generalization}

This model of coherence:
\begin{itemize}
    \item Does not require closure under logical entailment.
    \item Does not assign belief truth values or derive probabilities.
    \item Allows partial coherence in disconnected or contested substructures.
\end{itemize}

These distinctions mark a departure from both AGM-style belief revision~\cite{agm1985} and formal argumentation frameworks~\cite{hunter2020}, which are typically equipped with resolution or update procedures. Instead, our static formalism offers a neutral space for representing and interrogating belief tension, fragmentation, and misalignment—serving as a foundational layer for future dynamic or evaluative mechanisms.

% -------------- Section end marker --------------
%                _       _
%               ( )_    ( )
%    ___  _   _ | ,_)   | |__     __   _ __   __
%  /'___)( ) ( )| |     |  _ `\ /'__`\( '__)/'__`\
% ( (___ | (_) || |_    | | | |(  ___/| |  (  ___/
% `\____)`\___/'`\__)   (_) (_)`\____)(_)  `\____)
%
% -------------- Section end marker --------------

\section{Credibility and Confidence as Distinct Epistemic Dimensions}
\label{sec:credconf}

A foundational departure from probabilistic, logical, and argumentation-based belief models lies in our explicit separation between \emph{credibility} and \emph{confidence}. While most existing frameworks collapse epistemic strength into a single scalar (e.g., probability, weight, or score), our model disaggregates these into two orthogonal axes: one externally imposed and source-driven, the other internally emergent and structure-driven. This section clarifies the semantic and functional distinctions between these dimensions and articulates their joint expressive power.

\subsection*{Credibility: Source-Driven Evaluation}

The \emph{credibility} function
\[
\mathrm{cred}: N \rightarrow [0,1]
\]
assigns each belief node a value capturing the reliability, trustworthiness, or epistemic authority of its origin. It reflects assessments about the source of a belief, not the belief's integration into the larger system. 

\begin{itemize}
    \item \textbf{Input-facing}: Encodes trust in the origin or channel of belief acquisition.
    \item \textbf{Exogenous}: Defined independently of the system’s structural configuration.
    \item \textbf{Static}: Immutable within the present framework, barring manual reassignment.
\end{itemize}

\emph{Credibility} serves as an epistemic “prior” akin to source reliability in Bayesian epistemology or argument weighting in structured argumentation, but it is not used for propagation or inference. Unlike in Bayesian models~\cite{pearl1988probabilistic}, credibility here does not entail probabilistic coherence.

\subsection*{Confidence: Structure-Driven Evaluation}

The \emph{confidence} function
\[
\mathrm{conf}: N \rightarrow [0,1]
\]
assigns each belief node a score reflecting the degree of epistemic support it receives from other beliefs in the system. It encodes internal justification rather than external authority.

\begin{itemize}
    \item \textbf{Structure-facing}: Measures a belief’s embeddedness, reinforcement, or exposure to contradiction.
    \item \textbf{Endogenous}: Potentially computable from the topology and edge types of the belief graph.
    \item \textbf{Latently dynamic}: While fixed in the current model, it is amenable to future derivation and revision.
\end{itemize}

Conceptually, confidence generalizes internal justification from modal logic and extends structural support notions from argumentation frameworks~\cite{hunter2022epistemic}. Its future derivation could take the form:
\[
\mathrm{conf}(n_j) \propto \sum_{(n_i, n_j) \in E \text{ and } \mathrm{type}(n_i, n_j) = \text{support}} w(n_i, n_j) \cdot \mathrm{conf}(n_i)
\]
where $w(n_i, n_j)$ denotes the weight or strength of the epistemic support.

\subsection*{Interpretive Summary}

The dual assignment of $(\mathrm{cred}, \mathrm{conf})$ allows for epistemic states inaccessible to single-metric belief models:
\begin{itemize}
    \item \textbf{High-cred, low-conf}: Belief is trusted due to source, yet poorly supported internally.
    \item \textbf{Low-cred, high-conf}: Belief arises from a dubious source, but is internally reinforced.
    \item \textbf{Aligned}: Credibility and confidence reinforce each other—potentially signaling epistemic stability.
\end{itemize}

These configurations distinguish our framework from probabilistic, logical, or revision-based approaches, where such tension is typically resolved or suppressed.

\subsection*{Functional Independence}

In the present static formalism, we treat $\mathrm{cred}$ and $\mathrm{conf}$ as fully independent. Their values may contradict, align, or diverge without constraint. Unlike belief revision models (e.g., AGM~\cite{agm1985}) or dynamic consensus frameworks~\cite{bernardo2024bounded}, we impose no assumptions about how source reliability should influence structural confidence or vice versa.

\subsection*{Conceptual Benefit}

This separation directly addresses key limitations in prior models:
\begin{itemize}
    \item Probabilistic approaches conflate source trust and system integration into a single value.
    \item Logical and modal systems neglect scalar evaluations entirely or restrict them to binary modalities.
    \item Argumentation models often tie belief strength to network propagation, without distinguishing exogenous priors.
\end{itemize}

By disentangling the origin and integration of belief, our model enables the representation of:
\begin{itemize}
    \item \textbf{Epistemic misalignment}: e.g., beliefs that are credible but ungrounded, or grounded but incredible.
    \item \textbf{Justificatory tension}: zones where credibility and confidence diverge, inviting scrutiny or re-evaluation.
    \item \textbf{Heterogeneous landscapes}: systems where different beliefs exhibit structurally different epistemic profiles.
\end{itemize}

This distinction supports nuanced epistemic diagnostics without requiring deductive closure, inference rules, or belief updates—preserving representational flexibility for downstream interpretive or computational layers.

% -------------- Section end marker --------------
%                _       _
%               ( )_    ( )
%    ___  _   _ | ,_)   | |__     __   _ __   __
%  /'___)( ) ( )| |     |  _ `\ /'__`\( '__)/'__`\
% ( (___ | (_) || |_    | | | |(  ___/| |  (  ___/
% `\____)`\___/'`\__)   (_) (_)`\____)(_)  `\____)
%
% -------------- Section end marker --------------

%\begin{figure}[htbp]
   % \centering
  %  \includegraphics[width=0.48\textwidth]{figures/depth_sensitivity.png}
  %  \caption{Final MSE vs. depth for PGNN. Models remain stable up to 10 layers.}
   % \label{fig:depth-sensitivity}
%\end{figure}
% -------------- Section end marker --------------
%                _       _
%               ( )_    ( )
%    ___  _   _ | ,_)   | |__     __   _ __   __
%  /'___)( ) ( )| |     |  _ `\ /'__`\( '__)/'__`\
% ( (___ | (_) || |_    | | | |(  ___/| |  (  ___/
% `\____)`\___/'`\__)   (_) (_)`\____)(_)  `\____)
%
% -------------- Section end marker --------------

\section{Structural Interpretation and Epistemic Use Cases}
\label{sec:usecases}

Although the present framework is deliberately static—eschewing dynamics, updates, or inference—it offers substantial epistemic utility through its structural design. By modeling belief systems as graphs annotated with orthogonal evaluations of credibility and confidence, our framework enables diagnostic, interpretive, and preparatory applications unavailable to classical belief models. This section outlines several representative use cases, each emphasizing a distinct facet of the framework’s analytical potential.

\subsection*{Use Case 1: Identifying Epistemic Tension}

Classical belief representations either assume consistency (logic-based) or encode belief tension probabilistically. Our graph-based model permits direct localization of tension zones—subgraphs where contradiction or undermining structures appear.

\begin{itemize}
    \item \textbf{Example}: A highly credible belief node is targeted by multiple contradictory edges from structurally coherent but low-credibility nodes.
    \item \textbf{Interpretation}: Despite its trusted provenance, the belief is structurally embattled, flagging it as epistemically unstable.
\end{itemize}

This capability generalizes beyond the binary attack/support dichotomy in argumentation frameworks~\cite{dung1995,hunter2022epistemic}, by enabling scalar and multi-relational diagnostics.

\subsection*{Use Case 2: Isolating Coherent Subsystems}

Unlike probabilistic models that focus on global consistency, our model allows the extraction of locally coherent regions—subgraphs that are internally non-contradictory and mutually reinforcing.

\begin{itemize}
    \item \textbf{Example}: A cluster of beliefs with only support or qualification edges and internally consistent confidence.
    \item \textbf{Interpretation}: An epistemic “island” of stability that may serve as a candidate for reliable reasoning.
\end{itemize}

This structural modularity supports applications in belief filtering, policy grounding, and selective activation within larger, possibly inconsistent systems.

\subsection*{Use Case 3: Mapping Source–Structure Divergence}

Our model uniquely enables epistemic dual diagnostics by jointly analyzing credibility (source-based) and confidence (structure-based). This facilitates identification of belief nodes where the two metrics misalign.

\begin{itemize}
    \item \textbf{Example}: A belief originating from a weak or contested source attains high confidence due to extensive internal reinforcement.
    \item \textbf{Interpretation}: The belief has gained epistemic traction independently of its origin—a case invisible in source-only or structure-only frameworks.
\end{itemize}

Such divergence highlights structural belief emergence and rational over-endorsement, extending insights from analogical alignment models~\cite{olsson2024analogy}.

\subsection*{Use Case 4: Diagnostic Representation of Belief States}

In multi-agent or aggregated belief settings, the belief graph offers a snapshot diagnostic of structural organization without presupposing inference rules or truth-functional semantics.

\begin{itemize}
    \item \textbf{Example}: A large-scale belief map reveals distinct, weakly connected clusters corresponding to epistemic subcommunities or belief factions.
    \item \textbf{Interpretation}: Enables cross-agent comparison, coalition detection, or visualization of structural voids—tasks typically outside the scope of logical or probabilistic models.
\end{itemize}

This supports epistemic cartography in social, organizational, or multi-model AI contexts~\cite{carmina2022hubs,bernardo2024bounded}.

\subsection*{Use Case 5: Preprocessing for Reasoning Architectures}

While our model avoids inference, it can serve as a pre-filtering or scaffolding mechanism for downstream reasoning modules—logical, probabilistic, or hybrid.

\begin{itemize}
    \item \textbf{Example}: A maximally coherent subgraph is extracted and passed to a deductive system as axiomatic input.
    \item \textbf{Interpretation}: The belief graph guides principled, selective commitment to structurally well-supported propositions.
\end{itemize}

This aligns with current efforts to augment inference systems with structurally pre-conditioned inputs~\cite{hase2023methods},  ~\cite{Rocha2022}.

\subsection*{Remarks}

These use cases underscore the expressive and analytical distinctiveness of the proposed framework:
\begin{itemize}
    \item It generalizes belief evaluation beyond scalar inference (as in Bayesian or Dempster–Shafer models).
    \item It supports epistemic mapping without relying on attack/defense cycles (as in classical argumentation).
    \item It enables static epistemic diagnostics without enforcing closure or truth-functional rules.
\end{itemize}

These capacities suggest the belief graph model is not merely a representational artifact, but a general-purpose substrate upon which dynamic, inferential, or learning mechanisms may later operate.

\subsection*{Illustrative Example: Divergence Between Credibility and Confidence}

\begin{figure}[htbp]
    \centering
    \includegraphics[width=0.8\textwidth]{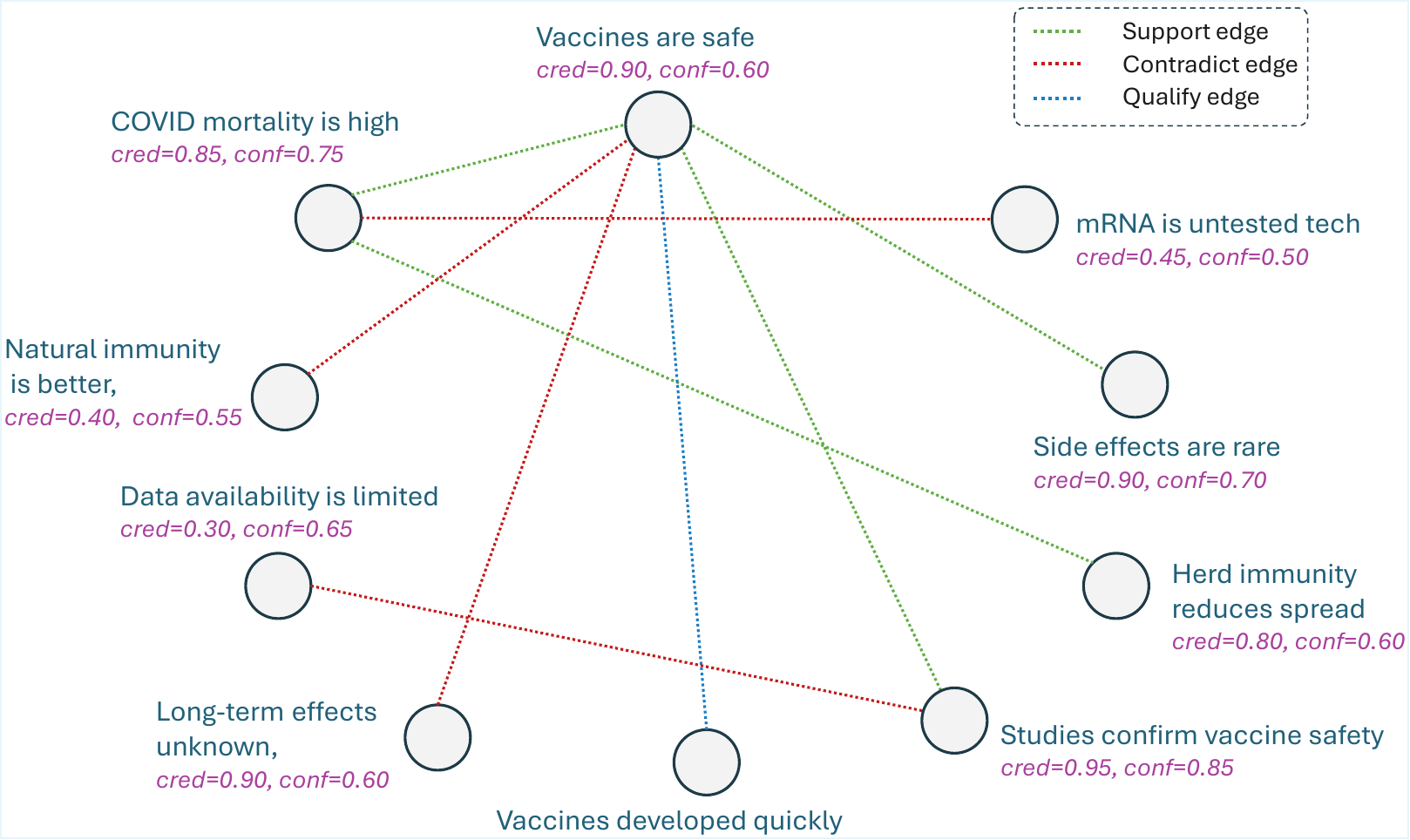}
    \caption{Belief graph illustrating divergence between external credibility and internal confidence. Nodes represent beliefs related to COVID-19 vaccination, annotated with both \emph{cred} (source reliability) and \emph{conf} (structural support). Edges encode epistemic relations—support or contradiction. The example highlights regions where credibility and confidence diverge, illustrating epistemic misalignment and structural tension.}
    \label{fig:belief_network_example}
\end{figure}

Figure~\ref{fig:belief_network_example} illustrates a real-world-inspired belief system centered on COVID-19 vaccine discourse. It visualizes how credibility and confidence can diverge—e.g., highly credible beliefs may receive limited structural support, while less credible beliefs may become internally reinforced. This example underscores the value of decoupling source trust from structural integration when analyzing epistemic configurations.

% -------------- Section end marker --------------
%                _       _
%               ( )_    ( )
%    ___  _   _ | ,_)   | |__     __   _ __   __
%  /'___)( ) ( )| |     |  _ `\ /'__`\( '__)/'__`\
% ( (___ | (_) || |_    | | | |(  ___/| |  (  ___/
% `\____)`\___/'`\__)   (_) (_)`\____)(_)  `\____)
%
% -------------- Section end marker --------------

% -------------- Section end marker --------------
%                _       _
%               ( )_    ( )
%    ___  _   _ | ,_)   | |__     __   _ __   __
%  /'___)( ) ( )| |     |  _ `\ /'__`\( '__)/'__`\
% ( (___ | (_) || |_    | | | |(  ___/| |  (  ___/
% `\____)`\___/'`\__)   (_) (_)`\____)(_)  `\____)
%
% -------------- Section end marker --------------
%\acresetall

\section{Conclusion and Outlook}
\label{sec:conclusion}

This paper introduced a static, structural framework for modeling belief systems as directed, weighted graphs. In contrast to traditional models grounded in logic, probability, or revision dynamics, our approach explicitly decouples two core epistemic dimensions: external \emph{credibility} (source trust) and internal \emph{confidence} (structural support). Epistemic relationships among beliefs are encoded as directed edges, optionally typed to distinguish reinforcement, qualification, or contradiction.

By refraining from inference procedures, belief revision rules, or probabilistic semantics, we isolate the structural organization of belief as a standalone modeling object. This enables the explicit representation of:
\begin{itemize}
    \item \textbf{Coherent substructures} within otherwise fragmented belief systems;
    \item \textbf{Zones of epistemic tension} arising from contradiction, misalignment, or unsupported claims;
    \item \textbf{Divergence between source and structure}, such as credible but unsupported beliefs, or coherent beliefs with dubious origins;
    \item \textbf{Diagnostic and interpretive applications} that remain inaccessible to scalar or deductively closed models.
\end{itemize}

Unlike argumentation frameworks or probabilistic reasoning systems, our model offers a domain-agnostic, non-inferential foundation for belief representation—suited for inspection, decomposition, and structural assessment, rather than resolution or conclusion derivation.

\subsection*{Outlook}

Several research directions emerge naturally from this foundational layer:

\begin{itemize}
    \item \textbf{Belief update and revision}: Defining mechanisms for node/edge reweighting, contradiction resolution, or structural restructuring in light of new information.
    \item \textbf{Confidence propagation algorithms}: Establishing principled ways to compute internal confidence from local support, contradiction, or redundancy patterns.
    \item \textbf{Multi-agent epistemics}: Modeling interaction between belief systems, including partial overlap, conflict detection, and belief exchange or fusion.
    \item \textbf{Integration with reasoning systems}: Using structurally coherent subgraphs as preconditioned inputs for logical, probabilistic, or neural inference modules.
    \item \textbf{Applications in epistemic mapping}: Employing belief graphs to visualize, segment, and compare belief states across communities, institutions, or AI systems.
\end{itemize}

These extensions are deliberately deferred. The present work lays a minimal structural groundwork—free from procedural assumptions—upon which diverse reasoning mechanisms, update dynamics, or learning architectures may later operate. By grounding belief modeling in graph-theoretic and dual-dimensional terms, we provide a versatile substrate for epistemic representation that remains robust in the face of contradiction, fragmentation, and incomplete information.

% -------------- Section end marker --------------
%                _       _
%               ( )_    ( )
%    ___  _   _ | ,_)   | |__     __   _ __   __
%  /'___)( ) ( )| |     |  _ `\ /'__`\( '__)/'__`\
% ( (___ | (_) || |_    | | | |(  ___/| |  (  ___/
% `\____)`\___/'`\__)   (_) (_)`\____)(_)  `\____)
%
% -------------- Section end marker --------------

%\bibliographystyle{IEEEtran}
%\bibliography{references}

%\nocite{*}
%\pagebreak[4]
\printbibliography
%\bibliography{main}
% Use to help equalize columns on last page
%\pagebreak[4]

\end{document}